\documentclass[sigconf]{acmart}

\AtBeginDocument{%
  \providecommand\BibTeX{{%
    \normalfont B\kern-0.5em{\scshape i\kern-0.25em b}\kern-0.8em\TeX}}}

\copyrightyear{2021}
\acmYear{2021}
\setcopyright{acmcopyright}
\acmConference[MM '21] {Proceedings of the 29th ACM International Conference on Multimedia}{October 20--24, 2021}{Virtual Event, China.}
\acmBooktitle{Proceedings of the 29th ACM Int'l Conference on Multimedia (MM '21), Oct. 20--24, 2021, Virtual Event, China}
\acmPrice{15.00}
\acmISBN{978-1-4503-8651-7/21/10}
\acmDOI{10.1145/3474085.3478560}

\begin{document}

\fancyhead{}
\title{ViDA-MAN: Visual Dialog with Digital Humans}



\author{Tong Shen$^{1}$, Jiawei Zuo$^{1}$, Fan Shi$^{1}$, Jin Zhang$^{2}$, Liqin Jiang$^{2}$,\\ Meng Chen$^{1}$, Zhengchen Zhang$^{1}$, Wei Zhang$^{1}$, Xiaodong He$^{1}$, Tao Mei$^{1}$}


\affiliation{%
  \institution{$^{1}$ JD AI Research, Beijing, China, \hspace{10pt}  $^{2}$ Migu Culture Technology, Beijing, China}
  \country{}}

\renewcommand{\shortauthors}{Tong Shen et al.}

\begin{abstract}
  We demonstrate ViDA-MAN, a digital-human agent for multi-modal interaction, which offers realtime audio-visual responses to instant speech inquiries. Compared to traditional text or voice-based system, ViDA-MAN offers human-like interactions (e.g, vivid voice, natural facial expression and body gestures). Given a speech request, the demonstration is able to response with high quality videos in sub-second latency.
  To deliver immersive user experience, ViDA-MAN seamlessly integrates multi-modal techniques including Acoustic Speech Recognition (ASR), multi-turn dialog, Text To Speech (TTS), talking heads video generation.
  Backed with large knowledge base, ViDA-MAN is able to chat with users on a number of topics including chit-chat, weather, device control, News recommendations, booking hotels, as well as answering questions via structured knowledge.
\end{abstract}




\maketitle

\vspace{-2mm}
\section{Introduction}
Digital humans are virtual avatars backed by Artificial Intelligence (AI), which are designed to behave like a real human. Agents powered by such systems can be applied in a wide range of scenarios such as personal assistant, customer service and News broadcasting. In this paper, we present ViDA-MAN, a multi-modal interaction system for digital humans. The system is complex by nature, integrating multimodal techniques such as ASR, TTS, dialog system, visual synthesis. 

One of the core parts of a digital human system is the ability to receive signals from the user and output the corresponding feedback, which can be viewed as a chatbot.
We develop a multi-type spoken dialogue system that can handle user requests by multiple dialog skills, such as chit-chat, task-oriented dialog, and question answering based on knowledge graph.

\begin{figure}[t]
  \centering
\begin{tabular}{ll}
 \includegraphics[width=0.45\linewidth]{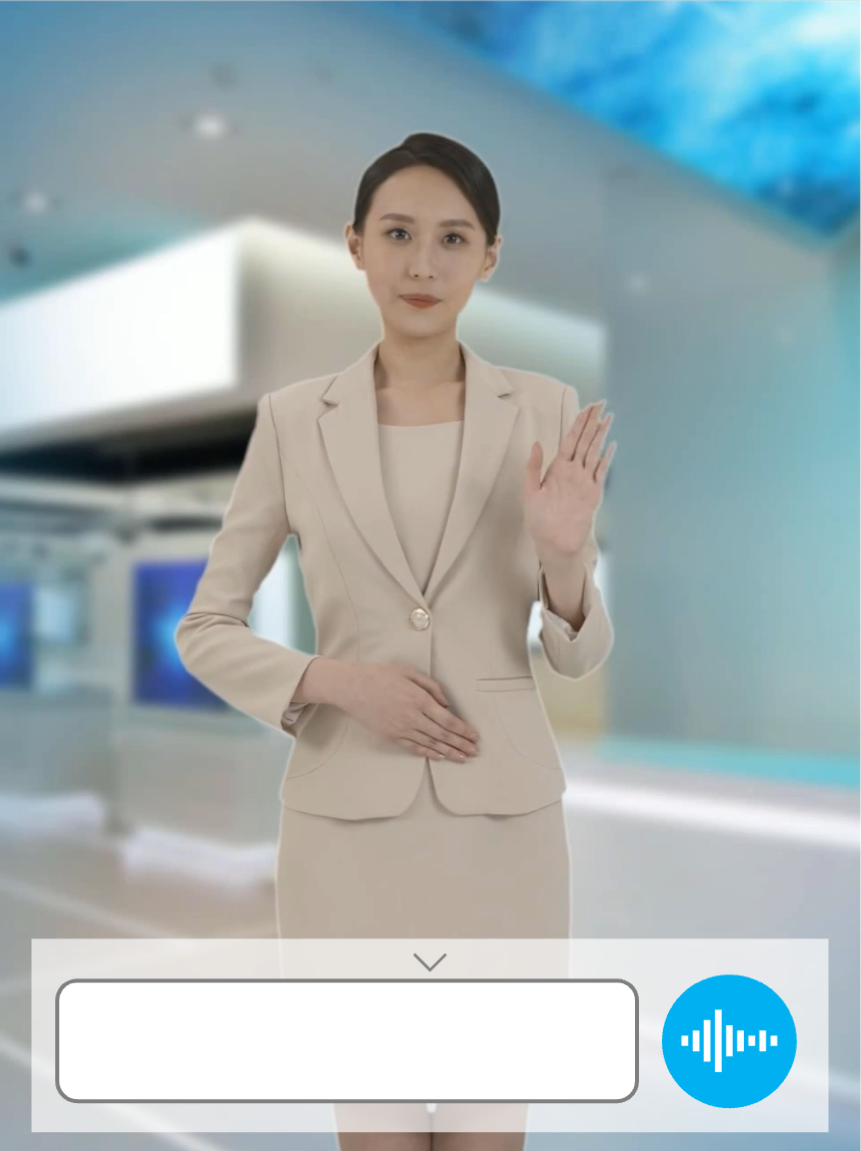} &
 \includegraphics[width=0.45\linewidth]{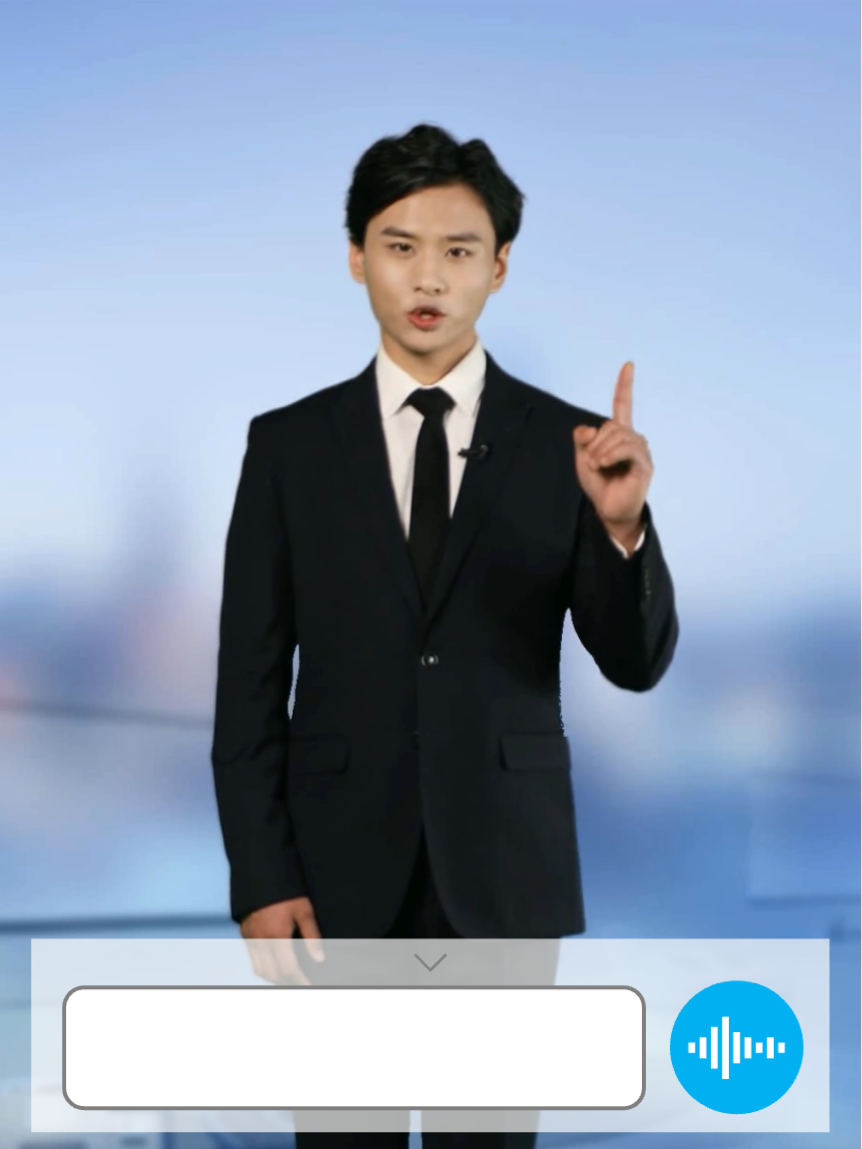} \\
\end{tabular}
\vspace{-2mm}
  \caption{Illustration of our interactive demo. The digital characters have realistic appearance, voice, natural facial expression and body motions, offering lifelike interaction experiences.}
  \label{fig_face_exp}
  \vspace{-2.5mm}
\end{figure}

What makes our ViDA-MAN different from a pure chatbot is its concrete visual appearance and voice, which expresses far more information than a pure text-based system, e.g. body language or facial expressions. The voice is empowered by a high quality TTS system, consisting a novel Duration Informed Auto-regressive Network (DIAN)\cite{song2021dian} and a speaker-specific neural vocoder LPCNet. 
The appearance is powered by neural rendering techniques \cite{Park_2019_CVPR, Neural_Point-Based_Graphics, Neural_Volumes, Deep_Appearance_Models_for_Face_Rendering, chan2019dance, v2v}. Different from a graphics rendering engines \cite{urengine,unity}, neural renderers do not require a specific high-quality 3D model and are able to produce far more realistic visual results. Figure \ref{fig_face_exp} demonstrates some examples.
In this paper, we present our digital human system, ViDA-MAN, to draw more attention on multi-modal interaction systems.


\vspace{-2mm}
\section{system architecture}
\begin{figure*}[t]
  \centering
  \includegraphics[width=0.94\linewidth]{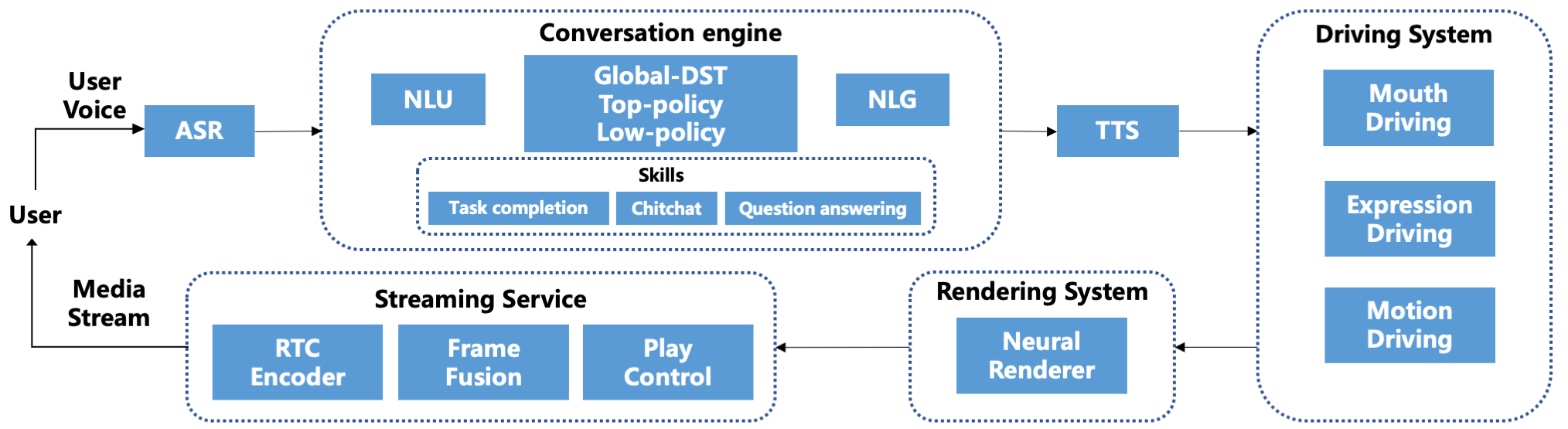}
  \vspace{-2mm}
  \caption{System architecture of ViDA-MAN. The system interacts with the user through human-like voice and appearances. Major sub-modules include ASR, Dialog, TTS, Driving $\&$ Visual Renderer, and Streaming service.}
  \label{fig_system}
  \vspace{-2mm}
\end{figure*}

The whole system is designed to pursue low latency and high visual fidelity, seeking intelligent and real-time interactions with a lifelike digital character. As shown in Figure \ref{fig_system}, the system mainly consists of six modules. The system accepts human voice by an ASR module and feeds it to a dialog system. The response is further translated to realistic voice using TTS. A driving system and a rendering system are responsible for updating the appearance. A streaming service is adopted to integrate everything and encode it into media stream back to the user.

\vspace{-2mm}
\subsection{Speech Module: ASR and TTS}
The speech module (ASR and TTS) acts as interpreters of the whole system. ASR has been widely studied for decades. We choose a widely-used HMM-DNN framework and use TDNN+LSTM \cite{Peddinti2015ATD} to implement the system. 
The TTS module is essential for human-like voice output. We adopt DIAN \cite{song2021dian} and a speaker-specific neural vocoder LPCNet \cite{Valin2019LPCNETIN}. 
By adopting the multi-speaker duration model and acoustic model with the speaker-dependent data, the TTS system is able to clone and generate the voice of every new speaker or speaking style, which greatly enriches diversity of choices under different scenarios and can better cooperate with other modules to present digital humans.

\vspace{-2mm}
\subsection{Dialog Module: Conversation Engine}

Our conversation engine is equipped with multiple dialog skills, which covers various domains including news, weather, movie, etc. As shown in Figure \ref{fig_system}, after the ASR, the transcriptions of the user's speech are firstly passed to a Natural Language Understanding (NLU) unit to identify the user’s intention and related entity information. The information is then formatted as the input to DST (Dialog State Tracker), which maintains the current state of the dialog. In order to coordinate different skills naturally, a hierarchical dialog policy controller \cite{kulkarni2016hierarchical} is adopted to select next dialogue actions. The main skills are controlled by top-level policy, whereas primitive actions are controlled by low-level policy. Finally, a model-based Natural Language Generator (NLG) \cite{wen2015semantically} is involved to convert agent actions to natural language responses.

\vspace{-2mm}
\subsection{Visual Module: Driving and Rendering}
The appearance of our digital human actually originates from our real human model. We first record videos of the human model speaking or making different poses. The footage is then used to reconstruct the person with intermediate representations. The role of the driving system is to produce desired representations given speech signals. Our driving system is mainly focused on the face region, i.e. making facial expressions, moving lips etc., but it can be extended to the whole body. The face is reconstructed and represented by a 3D Morphable Model (3DMM) \cite{3dmm, flame, bfm}.




The rendering system is mainly composed of a neural renderer, which converts the intermediate representation (3DMM mesh) into realistic images. We adopt a modified version of Vid2Vid \cite{v2v}, which is a video-to-video synthesis framework considering both spatial and temporal consistency. The model is trained by sequences of paired images, real face images and the reconstructed ones.

\vspace{-2mm}
\subsection{Realtime Streaming}
Our whole system is streaming-based, therefore the streaming service unit plays an essential role in controlling the pace, encoding the frames and broadcasting the media. There are three modules involved. A play control unit is used to control the general status of the digital human. Since our driving system is mainly related to facial movement, the body frames (actions, transitions, etc.) are pre-loaded and controlled by the unit. The render system only starts to render when there are facial movements, which saves computational cost and increases scalability. The face images are then processed and fused with the body frames through a frame fusion unit and further encoded by RTC (Real-Time Communication) encoder. The frames with the synced audio content are compressed into H.246 format and broadcasted by our own RTC platform.


\vspace{-2mm}
\section{System and User Interface}

Our dialog system is trained on multiple types of data including a knowledge graph with 180k triples, 22k entities, 20k QA pairs and million-scale NLU data, which covers topics such as celebrities, music, movies. ViDA-MAN supports tasks like enquiring weather, reserving hotels, searching news, etc. Moreover, it is also extendable to more domains in the future. 

Voice and appearance are two crucial components for providing human-like interactions. Our TTS benefits a lot from the novel framework and can generate high-quality voice. On a scale from 0 to 5, our TTS system reaches 4.2 MOS (mean opinion score) in our user study, which shows superior quality. Our rendering system also benefits from our joint 2D and 3D based generation networks that produce realistic talking faces in realtime. Figure \ref{fig_face_exp} shows examples of our user interface. 

ViDA-MAN supports streaming of 1080p video at 25 FPS (frames per second). The overall latency is around 400 milliseconds when properly deployed. One single GPU (Tesla P40) is able to support four instances simultaneously. 





%


\begin{acks}
This work was partially supported by the National Key R\&D Program of China under Grant No. 2020AAA0108600, Migu-JD Joint AI-Lab on Multimodal Interaction.
\end{acks}
\newpage


\bibliographystyle{ACM-Reference-Format}

\end{document}